\documentclass[lettersize,journal]{IEEEtran}
\usepackage{amsmath,amsfonts}
\usepackage{algorithmic}
\usepackage{algorithm}
\usepackage{array}
\usepackage[caption=false,font=normalsize,labelfont=sf,textfont=sf]{subfig}
\usepackage{textcomp}
\usepackage{stfloats}
\usepackage{url}
\usepackage{verbatim}
\usepackage{graphicx}
\usepackage{cite}
\usepackage{multirow}
\begin{document}

\title{DAgger Diffusion Navigation: DAgger Boosted \\Diffusion Policy for Vision-Language Navigation}

\author{Haoxiang Shi, 
Xiang Deng, 
Zaijing Li,
Gongwei Chen,
Yaowei Wang \IEEEmembership{, Member, IEEE}, \\
Liqiang Nie \IEEEmembership{, Senior Member, IEEE}




}

\markboth{Journal of \LaTeX\ Class Files,~Vol.~14, No.~8, August~2021}%
{Shell \MakeLowercase{\textit{et al.}}: A Sample Article Using IEEEtran.cls for IEEE Journals}

\IEEEpubid{}

\maketitle

\begin{abstract}

Vision-Language Navigation in Continuous Environments (VLN-CE) requires agents to follow natural language instructions through free-form 3D spaces. 
Existing VLN-CE approaches typically use a two-stage waypoint planning framework, where a high-level waypoint predictor generates the navigable waypoints, and then a navigation planner suggests the intermediate goals in the high-level action space.
However, this two-stage decomposition framework suffers from: (1) global sub-optimization due to the proxy objective in each stage, and 
(2) a performance bottleneck caused by the strong reliance on the quality of the first-stage predicted waypoints. 
To address these limitations, we propose DAgger Diffusion Navigation (DifNav), an end-to-end optimized VLN-CE policy that unifies the traditional two stages, i.e. waypoint generation and planning, into a single diffusion policy. 
Notably, DifNav employs a conditional diffusion policy to directly model multi-modal action distributions over future actions in continuous navigation space, eliminating the need for a waypoint predictor while enabling the agent to capture multiple possible instruction-following behaviors.
To address the issues of compounding error in imitation learning and enhance spatial reasoning in long-horizon navigation tasks, we employ DAgger for online policy training and expert trajectory augmentation, and use the aggregated data to further fine-tune the policy. This approach significantly improves the policy's robustness and its ability to recover from error states.
Extensive experiments on benchmark datasets demonstrate that, even without a waypoint predictor, the proposed method substantially outperforms previous state-of-the-art two-stage waypoint-based models in terms of navigation performance.
Our code is available at: \url{https://github.com/Tokishx/DifNav/}.
\end{abstract}


\section{Introduction}
\IEEEPARstart{V}ision-and-Language Navigation (VLN)\cite{anderson2018vision} tasks require an embodied agent to follow natural language instructions to reach the target location in complex environments. Many studies have been conducted on this task in the past few years \cite{wang2019reinforced,hong2021vln,chen2022think,zhou2024navgpt}. However, most of them have focused on discrete environments, where the agent's location and target are restricted to a fixed set of waypoints within a predefined map, significantly limiting the agent's exploration space. 
To align the VLN settings more closely with real-world navigation environments, Krantz et al. \cite{krantz2020beyond} introduced the Vision-and-Language Navigation in continuous environment (VLN-CE) setting and simulator, which enables agents to perform low-level actions (e.g., step forward 0.15m, turn right/left 15°) within realistic scenes. The handling of VLN-CE is challenging because the agent must follow high-level language instructions while executing a sequence of fine-grained navigational decisions (average 55 actions per episode) in a partially observable environment. As a result, performance in VLN-CE is significantly poorer compared to discrete VLN settings\cite{krantz2022sim}.

\begin{figure}[!t]
\centering
\includegraphics[width=3.4in]{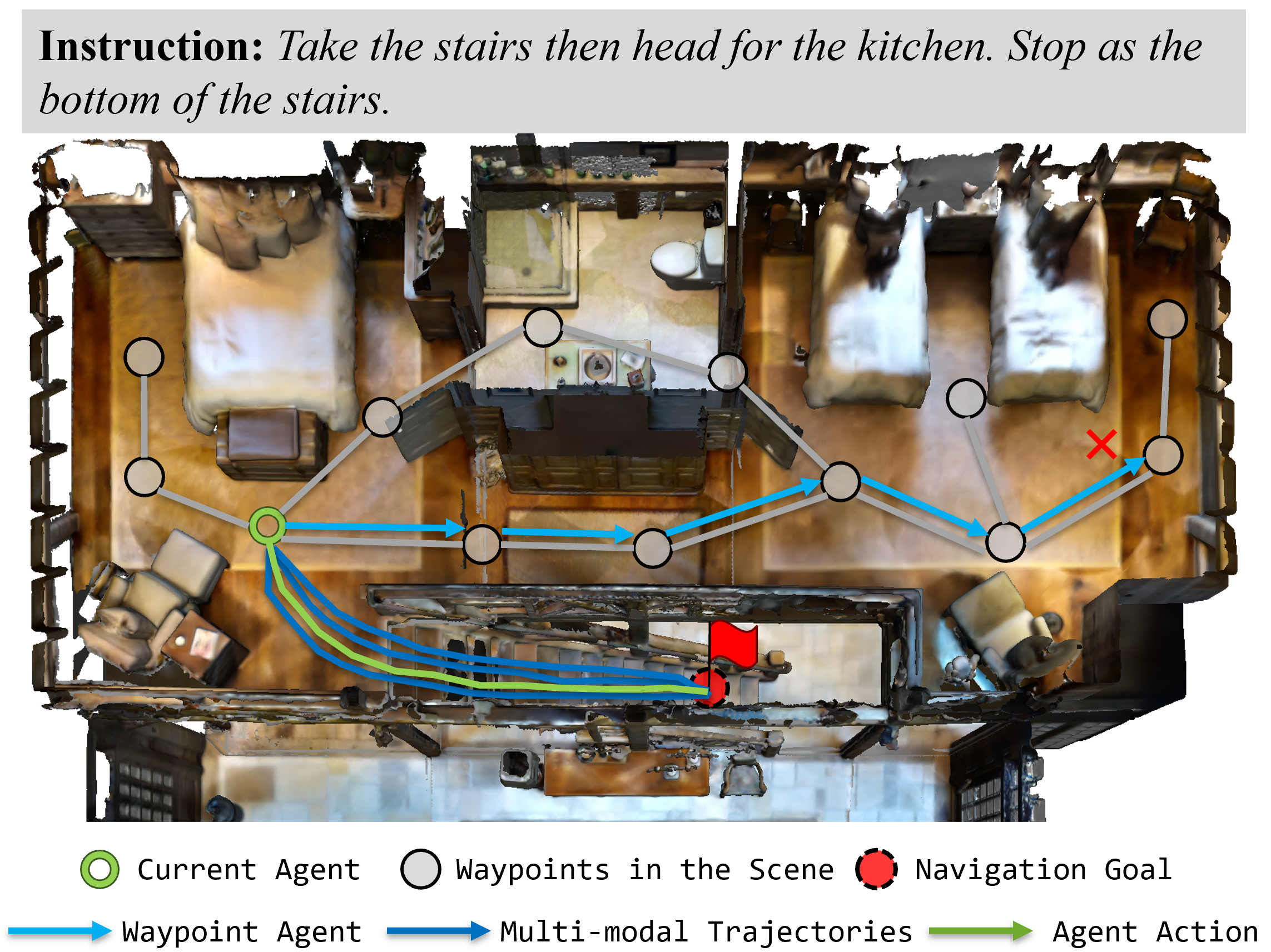}
\caption{Comparison of two-stage waypoint-based approaches (top) and DifNav (down). The two-stage waypoint-based approaches rely on a trained waypoint predictor to generate navigable waypoints and select the most suitable action for navigation. However, if the waypoint predictor fail to generate the correct waypoint towards destination, the navigation will fail. In contrast, DifNav directly models the multi-modal action distribution in navigation space and samples executable actions that follow instructions.}
\label{fig:introducation}
\end{figure}

To bridge the gap between the VLN task in the discrete environment and that in the continuous environment, conventional VLN-CE approaches adopt a two-stage waypoint-based framework that decouples the task into candidate waypoint generation and high-level sub-goal planning\cite{shi2025smartway}. These approaches first employ a trained waypoint predictor to generate several navigable waypoints around the agent, effectively transforming the continuous action space into a discrete one. 
Next, text-to-image grounding, which is commonly used in discrete VLN approaches, is used to select the most appropriate sub-goal from the generated waypoints based on the navigation instructions\cite{hong2022bridging}. 
This two-stage framework has significantly improved performance on VLN-CE tasks compared to low-level policies\cite{wang2023gridmm,wang2024lookahead,zhou2024navgpt2}. 

However, this framework suffers from global sub-optimality due to the use of different proxy objectives in each stage: the waypoint predictor is trained to generate accessible waypoints based on the environment's connectivity graph, and the sub-goal planner is optimized for the overall navigation task.
Additionally, the sub-goal planner is trained on an action space that depends on the generated waypoints, which can lead to error propagation between stages.
As shown in Fig. \ref{fig:introducation}, if the waypoint predictor produces misleading waypoints, such as those that are inaccessible or do not lead to the goal, the planner may generate sub-goals that lead the agent to an incorrect location.
Due to the limited action space defined by the generated waypoints, it becomes nearly impossible for the agent to recover from such errors.
Recent work has attempted to alleviate this issue by enhancing the spatial reasoning capabilities of the waypoint predictor \cite{zhang2024narrowing,shi2025smartway}. However, these solutions increase model complexity and only partly address the fundamental problem, namely that the agent remains vulnerable to suboptimal intermediate waypoint predictions.

An appealing alternative is an end-to-end approach that directly maps visual observations and language instructions to navigation actions, without relying on explicit intermediate waypoints. In principle, such a policy can learn to optimize for the final goal at each step, thereby potentially avoiding the misalignment issues introduced by waypoint proposals.
Previous VLN agents in discrete graph-based environments have successfully leveraged end-to-end learning via Behavioral Cloning (BC)\cite{wang2023scaling,chen2022think,li2023kerm}, a classical imitation learning approach. However, such methods have shown limited success in continuous settings, often underperforming compared to their discrete counterparts\cite{krantz2020beyond, krantz2021waypoint, chen2022weakly}. 
This performance gap arises from two key challenges. First, long-horizon decision making, coupled with the need to infer spatial accessibility for effective obstacle avoidance, is inherently difficult in continuous environments. Second, policies trained via BC from expert demonstrations suffer from severe compounding error due to the distributional shift between training and testing. Previous work has addressed this issue by introducing 'student-forcing' to reduce distribution mismatch. Other approaches have explored collecting extensive online data or incorporating reinforcement learning to mitigate the limitations of BC\cite{krantz2021waypoint, zhang2024navid}. However, these methods require massive training and still fall short of the performance achieved by approaches based on two-stage frameworks.

\begin{figure}[!t]
\centering
\includegraphics[width=3.4in]{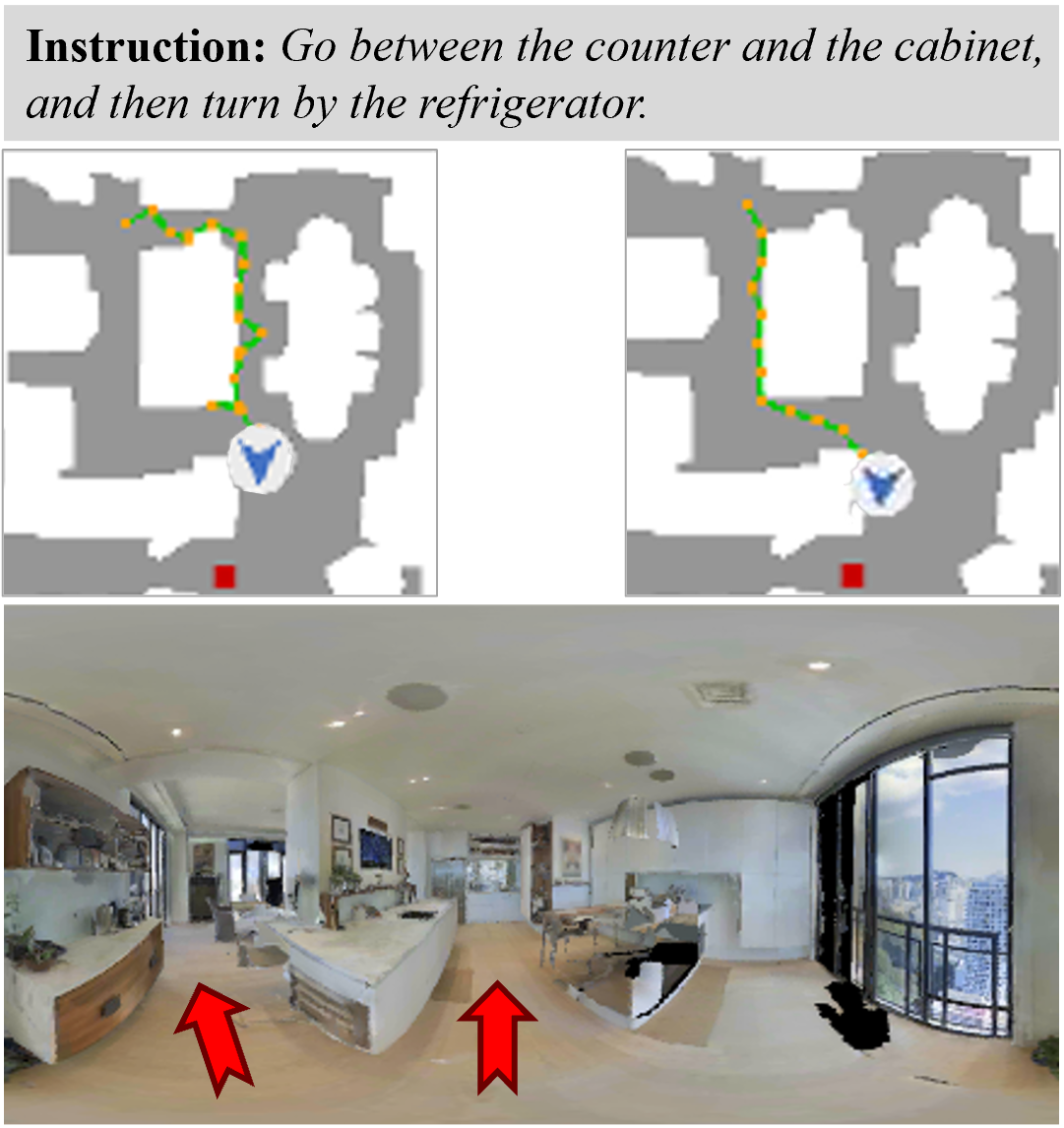}
\caption{\textbf{Multiple plausible trajectories corresponding to a single navigation instruction.}
The instruction allows for semantically equivalent paths that lead to the same goal, due to scene layout and linguistic ambiguity.}
\label{fig:multi-modal}
\end{figure}

Inspired by the success of diffusion policies in robotic manipulation, where policies are modeled as denoising diffusion processes over the action space \cite{chi2023diffusion}, we extend this paradigm to VLN. In this paper, we propose \textbf{D}Agger-boosted D\textbf{if}fusion \textbf{Nav}igation (DifNav), an end-to-end diffusion policy enhanced by Dataset Aggregation (DAgger) \cite{ross2011reduction}. DifNav directly models the multi-modal action distribution in continuous navigation space and samples instruction-conditioned actions from this distribution.
Specifically, DifNav represents the agent’s action as the output of a diffusion policy conditioned on the navigation instruction and visual observations. This formulation enables the policy to model multi-modal action distributions more effectively than traditional discriminative classification or regression-based methods, which are typically restricted to single-modal predictions.
As shown in Fig. \ref{fig:introducation}, without relying on the predicted waypoints,  DifNav models a multi-modal action distribution at each decision step and samples the executable action from it.
In VLN-CE, multi-modality in action decision-making is crucial because there are often multiple equivalent paths or ambiguous language that could imply different correct actions, as illustrated in Fig. \ref{fig:multi-modal}. By training a diffusion policy, our agent can naturally capture such ambiguity, producing a rich distribution over possible actions rather than committing to a single deterministic output. 
Furthermore, owing to the training stability and expressiveness in high-dimensional action spaces \cite{zhang2024effective}, the diffusion policy is well-suited for the joint modeling of cross-modal grounding and spatial reasoning, which is essential in end-to-end training for VLN-CE policies.

To address the compounding errors commonly observed in end-to-end models trained via BC, we enhance policy training using DAgger, an interactive imitation learning algorithm. DAgger augments the distribution of training data by executing the current policy and querying the expert for corrective actions whenever the agent deviates from the expert trajectory.
DAgger has been widely adopted to mitigate exposure bias in previous VLN-CE approaches\cite{an2024etpnav,an2023bevbert}. However, existing work has largely overlooked the multi-modal nature of demonstration trajectories generated during DAgger training, which can lead to policy inconsistency and degraded performance. In contrast, our method leverages the strong capability of diffusion policy to model multi-modal action distributions\cite{lee2024diffdagger}, effectively capturing diverse expert behaviors and achieving better alignment with the DAgger-augmented data distribution.
Following DAgger-boosted training, DifNav demonstrates the ability to recover from error states, resulting in significantly improved robustness and obstacle avoidance performance.

We conduct experiments on the R2R-CE benchmark\cite{krantz2020beyond}. The experimental results demonstrate that DifNav significantly outperforms state-of-the-art methods, despite not relying on a waypoint predictor.

The main contributions of our work are summarized as follows: 

\begin{itemize}
\item Different from two-stage waypoint-based approaches, we propose an end-to-end policy that directly predicts navigation actions using a diffusion policy. To the best of our knowledge, DifNav is the first diffusion-based policy introduced for vision-and-language navigation. 
Our end-to-end model leverages a conditional diffusion process to model the multi-modal action distribution, and samples from this distribution to generate the next action.

\item We propose a DAgger-boosted training regime for DifNav to address the compounding errors commonly encountered in BC. By aggregating expert corrections on the fly, the policy learns to recover from incorrect states, significantly improving robustness and obstacle avoidance capabilities.

\item Extensive experiments on the R2R-CE benchmark demonstrate that DifNav effectively models multi-modal action distributions for the VLN-CE task and consistently outperforms existing two-stage waypoint-based approaches in navigation performance.
\end{itemize}

\section{Related Work}
\subsection{Vision-and-Language Navigation}
Vision-and-language navigation, which grounds linguistic instructions in perceptions to facilitate goal-directed actions, plays a pivotal role in embodied intelligence. 
Several navigation datasets have been introduced to support VLN tasks, particularly for indoor navigation with step-by-step instructions, such as R2R \cite{anderson2018vision}, R4R \cite{sotp2019acl}, RxR\cite{deitke2022retrospectives}.

The early VLN methods use sequence-to-sequence LSTMs with cross-modal attention to predict the next actions \cite{ma2019self}. Subsequently, transformer-based architectures have demonstrated strong performance in VLN tasks due to their ability to generate rich multi-modality representations \cite{wang2022less}.  This framework has been further extended through the use of topological maps \cite{chen2022think} or metric maps \cite{wang2023gridmm} as memory structures to model action sequences more effectively.
Recently, with the rapid advancement of Large Language Models (LLMs)\cite{touvron2023llama} and Vision-Language Models (VLMs)\cite{liu2023visual}, large models have emerged as a promising approach to improve navigation performance \cite{chen2024mapgpt}, largely due to the open-world knowledge acquired through large-scale pretraining.

\subsection{Vision Language Navigation in Continuous Environment}
In order to enable agents to act more realistically, simulators for continuous environments such as AI2-THOR \cite{kolve2017ai2}, Gibson \cite{xia2018gibson}, and Habitat \cite{savva2019habitat} have been widely developed for embodied AI research.
To study vision-and-language navigation in continuous environments (VLN-CE), Krantz et al. \cite{krantz2020beyond} transfer the discrete paths in the R2R and RxR datasets to continuous trajectories based on the Habitat-MP3D simulator.

Without the pre-defined connectivity graph available in discrete environments, a VLN-CE agent must navigate by following instructions while simultaneously inferring spatial accessibility to avoid obstacles, which makes the task significantly more challenging due to the longer navigation horizon. 
This difficulty results in approximately a 20\% gap in the success rate for agents with the same architecture when transitioning from discrete to continuous environments\cite{krantz2020beyond}.
To bridge this gap, modular waypoint-based approaches have emerged as a popular solution\cite{Hong_2021_CVPR, krantz2022sim, lin2024correctable}.
These methods decompose the VLN-CE task into candidate waypoint generation and high-level sub-goal planning. First, a trained waypoint predictor is used to generate accessible waypoints, effectively forming a high-level action space similar to that of discrete environments. The navigation task can then be tackled using existing discrete VLN methods.
To train a waypoint predictor capable of generating accessible waypoints, CWP \cite{hong2022bridging} first constructs a navigable connectivity map in VLN-CE by refining the connectivity map from the discrete VLN environment, and then uses the resulting nodes as ground-truth waypoints for training. Zhang et al. \cite{zhang2024narrowing} use the obstacle semantic map to mask RGB-D perceptions, preventing the generation of waypoints within obstructed areas. 
Wang et al. \cite{wang2024lookahead} encode both semantic and occupancy maps to predict a traversable map in real-world environments, and then select the top-K waypoints with the highest probability scores in the map as candidate navigation targets. AO-Planner \cite{chen2025affordances} leverages grounded SAM \cite{ren2024grounded} to identify navigational affordance regions (e.g., floors) to generate navigable waypoints. 
However, trained waypoint predictors can produce suboptimal waypoints, introducing a mismatch between the training and deployment domains, which ultimately limits the agent's navigation performance.

\subsection{Diffusion Policy}
Diffusion models are generative models that learn to gradually transform noise into samples from a target distribution. They have demonstrated remarkable capability in modeling complex, high-dimensional data distributions in various vision tasks\cite{rombach2022high}. 
Recently, diffusion models have been widely adopted as policy representations in robotic manipulation tasks \cite{ma2024hierarchical, hao2024language}, due to their advantages in stable training, accurate modeling of multi-modal action distributions, and effectiveness in high-dimensional output spaces\cite{chi2023diffusion}.

In vision navigation, NoMaD \cite{sridhar2024nomad} employs a diffusion policy to model the multi-modal action distribution for short-term exploration and image-goal navigation in fully observable environments. LDP \cite{yu2024ldp} utilizes reinforcement learning to large-scale collect expert data, and incorporates global paths as additional conditions for navigation and obstacle avoidance. Zhang et al. \cite{zhang2024versatile} further investigate value-guided diffusion policies for navigation in partially observable environments by adopting QMDP as the reward function. 
However, the use of diffusion policy for instruction-guided navigation remains underexplored. In this work, we introduce the navigation instruction as an additional condition input to enable instruction-grounded navigation.
To address compounding errors typically caused by Behavioral Cloning (BC), we employ an interactive demonstrator inspired by DAgger to augment the training data distribution through online interaction. This approach substantially improves the robustness and obstacle avoidance performance of the policy.

\section{Method}
\subsection{Task setup}
We focus on the task of vision-and-language navigation in continuous environments (VLN-CE) \cite{krantz2020beyond}, a practical setting in which an agent navigates through a mesh-based 3D environment using low-level actions, guided by natural language instructions.
The action space consists of a set of basic navigation actions, such as FORWARD (0.25m), ROTATE LEFT/RIGHT (15°), and STOP. VLN-CE is implemented using the Habitat Simulator\cite{savva2019habitat}, which renders environmental observations based on the Matterport3D scene dataset\cite{chang2017matterport3d}.
For each navigation episode, the agent receives a natural language instruction that serves as the navigation goal. We denote the instruction embedding, consisting of $L$ words, as $W =\{\hat{w}_i\}^{L}_{i=1}$. At each decision step, the agent receives panoramic RGB-D observations $O=\{I_{rgb}, I_d\}$  composed of 12 RGB images and 12 depth images captured from discrete horizontal viewpoints.

\subsection{Approach}

\begin{figure*}[!t]
\centering
\includegraphics[width=6.5in]{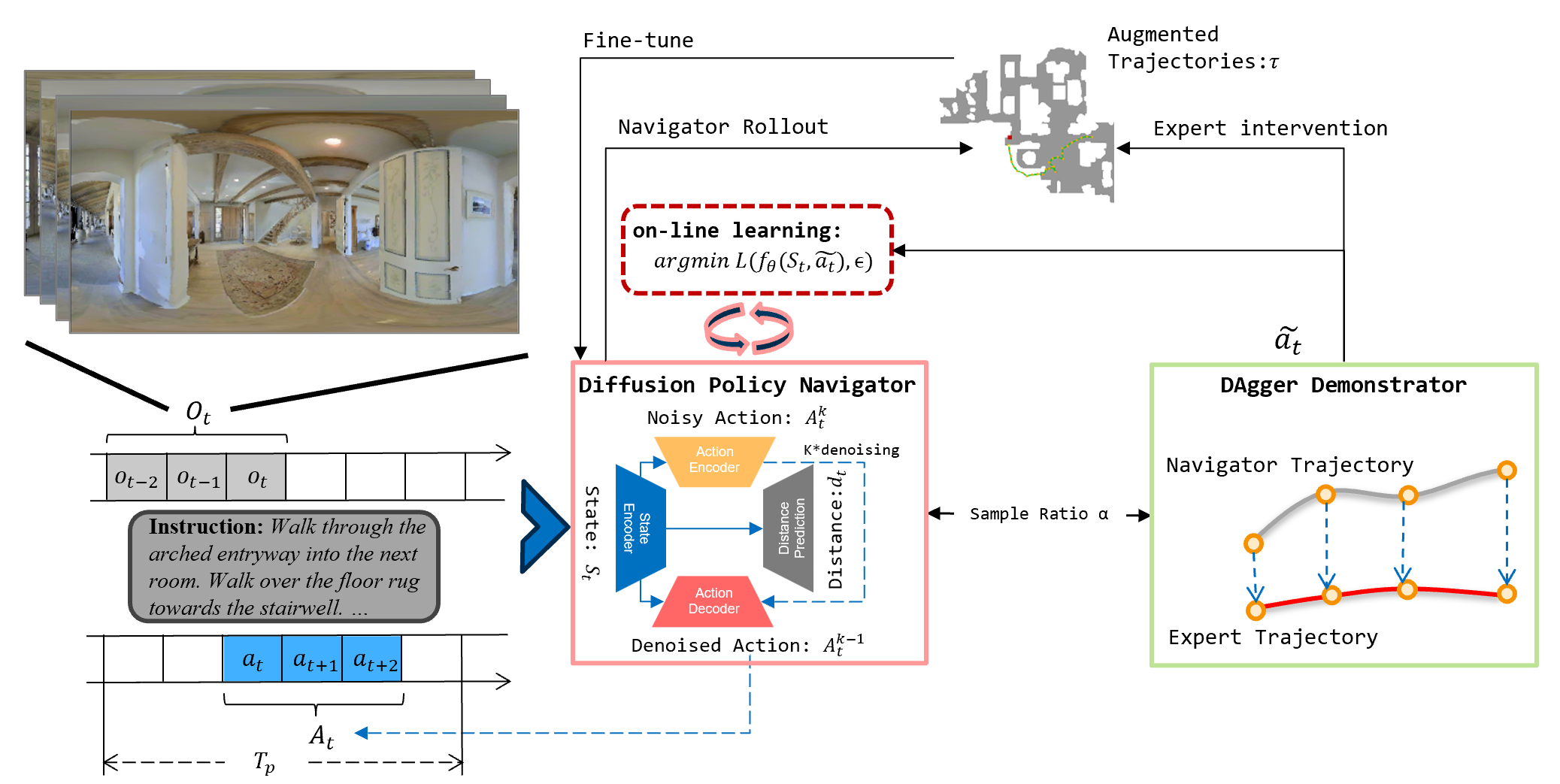}
\caption{\textbf{Model Architecture.} The DifNav framework first trains a Diffusion Policy Navigator, where a state encoder integrates historical visual observations and navigation instructions to form a latent state representation. This state is used to jointly train a diffusion policy for future action generation and a temporal distance predictor. After pretraining the navigator, we employ DAgger for online policy refinement. Through expert interventions provided by the DAgger Demonstrator, the policy is updated online, and the training data distribution is augmented. The augmented data are subsequently used for fine-tuning the policy.}
\label{fig:framework}
\end{figure*}

We propose an end-to-end VLN-CE policy, DAgger-boosted Diffusion Navigation (DifNav), which directly maps the state of the agent to actions in continuous navigation space. As illustrated in Fig. \ref{fig:framework}, the architecture of DifNav consists of three main components: a cross-modal state encoder, a conditional diffusion policy, and a temporal distance predictor. 
The state encoder encodes the current state of the agent by integrating information from its history trajectory, navigation instruction, and current observation.
The conditional diffusion policy models a multi-modal action distribution conditioned on the encoded state, enabling the sampling of executable and instruction-grounded actions.
The temporal distance predictor estimates the normalized distance from the agent to the navigation goal.

To address the compounding errors inherent in Behavior Cloning (BC), we employ DAgger-boosted online training to refine the policy and augment the training data distribution for subsequent fine-tuning.

The paper proceeds by introducing the cross-modal state encoder in \ref{State Encoding}, the state-conditioned diffusion policy to model multi-modal action distribution in \ref{Multi-Modal Action Distribution Modeling}, the temporal distance predictor in \ref{Temporal Distance Predictor}, DAgger-boosted online training in \ref{DAgger-Boosted Online Learning}, and training details in \ref{Training Details}.

\subsubsection{State Encoding} \label{State Encoding}
Vision Encoding: At each decision step, given the panoramic RGB-D observation $O_t=\{I_{rgb},I_{d}\}$, we use the pre-trained visual encoder ViT/32 to extract RGB features $V^{rgb}_t$ and a pre-trained depth encoder trained on point-goal navigation tasks\cite{Wijmans2020DD-PPO} to obtain depth features $V^{depth}_t$, respectively. To capture the positional relationship between panoramic views, we encode the heading angle as $V_{t}^{angle}=(sin\theta,cos\theta)$, where $\theta$ represents the heading angle of each view. The final observation embeddings are constructed as follows:
\begin{equation}
\label{equa:state}
O_t = LN(W_VV^{rgb}_t) + LN(W_DV^{depth}_t) + LN(W_LV_{t}^{angle})
\end{equation}
where $LN$ denotes the layer normalization, and $W_V, W_D, W_L$ denote the learnable weight matrices in the linear layers. The observation embeddings are then passed through a panorama encoder. The panorama encoder uses a multilayer transformer to perform interview interaction and produce the final observation embeddings $O'_t$.

Instruction Encoding: For instruction encoding, each word embedding in $W$ is added with a position embedding and a token type embedding. All tokens are then fed into a pre-trained multilayer transformer to obtain word representations, denoted as $\widehat{W} =\{\hat{w}_i\}^L_{i=1}$.

History Trajectory Encoding: we use the average pooling of the visual representations $Avg(O'_t)$ to represent the waypoint at each decision step $t$. For the current agent position and a historical waypoint at time step $t$, we define the Euclidean distance as $d_t$, the trajectory distance as $t_t$, the relative heading angles as $h_t=(\sin\Theta, \cos\Theta)$, and the navigation step embedding as $u_t$. The history trajectory embeddings $T_t$ are constructed as follows, where $W_t$ denotes the learnable weight matrix in the linear layer:
 \begin{equation}
\label{epua:his-encode}
T_t=[\{Avg(O'_t)+LN(W_t[d_t;h_t])+u_t\}^t_{i=1}]
\end{equation}

Cross-Modal Reasoning: A multilayer cross-modal transformer is used to model both vision-language relationships and spatio-temporal dependencies. Each transformer layer consists of a cross-attention module followed by a self-attention module. In the cross-attention layer, the navigation trajectory $T_t$ serves as the query to attend to the encoded instruction embeddings $\widehat{W}$. The self-attention layer then takes encoded panoramic observation and navigation trajectory as inputs for action reasoning. The resulting output represents the agent's current state and is denoted as $S_t$.

\subsubsection{Multi-Modal Action Distribution Modeling} \label{Multi-Modal Action Distribution Modeling}
Different from previous approaches that directly predict actions from a predefined action space, our method models a multi-modal action distribution at each decision step and samples an executable, instruction-conditioned action from this distribution.
Specifically, we use a state-conditioned diffusion policy to model the multi-modal action distribution $p(a_t|S_t)$, where $a_t$ denotes the next navigation action.

To train the noise prediction network in the diffusion policy, we randomly sample a sequence of future actions $a^K_t$ from a Gaussian distribution as stochastic noise and perform $K$ iterations of denoising to generate a series of intermediate action sequences with progressively reduced noise $\{a^{K}_t,a^{K-1}_t,...,a^{1}_t,a^{0}_t\}$, ultimately obtaining the desired noise-free output$a^0_t$. The iterative denoising process follows the equation:
 \begin{equation}
\label{equa:denoise}
a^{k-1}_t=\alpha\cdot(a^K_t - \gamma\epsilon_{\theta}(S_t, a^k_t,k)+N(0,\sigma^2I))
\end{equation}
where $k$ is the current denoising step, $\epsilon_{\theta}$ is the noise prediction network paramterized by $\theta$, $\alpha, \gamma, \sigma$ are the hyperparameters of the noise scheduler.

The noise prediction network is conditioned on the current state of the agent $S_t$, which captures the cross-modal representation derived from historical observations and the navigation instruction. During training, we add noise to the ground-truth action and train $\epsilon_\theta$ to predict this noise. The predicted noise is then compared to the actual noise using a mean squared error (MSE) loss.
 \begin{equation}
\label{equa:dp-loss}
L_{WP}=MSE(\epsilon^k,(S_t,\epsilon_\theta(a_0+\epsilon^k,k)))
\end{equation}

\subsubsection{Temporal Distance Predictor} \label{Temporal Distance Predictor}

Predicting the distance between the agent and the goal is valuable for monitoring navigation progress and determining whether the goal has been reached. We use the current state of the agent $S_t$ as input to a multilayer perceptron network (MLP) $f_d(S_t)$ to predict the distance to the goal. This distance prediction network is trained jointly with the diffusion policy, using a weighting factor $\lambda$.
 \begin{equation}
\label{equa:distance}
L_{Dist}=\lambda\cdot MSE(d(o_t,g_t),f_d(S_t))
\end{equation}
where $d(o_t,g_t)$ denotes the temporal distance between the agent and the goal.

\subsubsection{DAgger-Boosted Online Learning} \label{DAgger-Boosted Online Learning}
To mitigate the compounding errors in BC, we employ the DAgger algorithm to perform online policy training within the Habitat Simulator. The core procedure of DAgger is summarized as follows:
\begin{equation}
\label{equa:DAgger}
\left\{ 
\begin{array}{ll}
a_t=\text{Rollout}(S_t) & p>\alpha \\ 
a_t=\text{label}_t & p \le \alpha
\end{array}{}
\right.
\end{equation}
where $\text{Rollout}(S_t)$ denotes the action predicted by the current policy given state $S_t$, $\text{label}_t$ denotes the expert action at time step $t$, and $\alpha$ is the probability threshold controlling the frequency of expert intervention. Specifically, we employ an interactive demonstrator to determine the expert action corresponding to the current state of the agent. At each decision step, the agent executes its own policy with probability $p$; otherwise, it queries the demonstrator for expert intervention. The retrieved expert action is also used to compute the denoising loss, as equation \ref{equa:denoise}.
During DAgger training, the demonstrator selects, within a predefined radius, the action that minimizes the geodesic distance to the expert trajectory, and assigns it as the expert demonstration. 

After each navigation episode, the resulting trajectory is collected to augment the training data distribution, and the losses from all decision steps are aggregated to update the policy.
To iteratively optimize the policy and expand the expert demonstrations, we perform multiple rounds of DAgger training. In our experiment, we conducted five rounds of DAgger training.

\subsubsection{Training Details} \label{Training Details}
We first use the low-level expert demonstrations in the R2R-CE dataset to train a default policy for online learning.
These demonstrations have a waypoint spacing of 0.25 meter, constrained by the action space. 
Training directly on this data causes the policy to replicate the same fine-grained spacing observed in the demonstration trajectories.
To expand the action generation space, we extract waypoints at intervals of $n$ from the demonstration trajectories, creating sparse trajectories with a waypoint spacing of $n\times0.25$ meters. This expanded action space is consistent with the typical range used by waypoint predictors in the VLN-CE task.
Since navigation trajectories in the R2R-CE dataset are often long (approximately 55 steps per episode), using all past observations as history is redundant. Therefore, we encode only the three most recent observations as the history trajectory during training.
For the diffusion policy output, we set the length of the generated action sequence to 1, which means that the model predicts only the next instruction-conditioned action at each step. DifNav is trained end-to-end with supervised learning using the following loss function:
 \begin{equation}
\label{equa:train-loss}
L_{DifNav}(\varphi,\theta,f_d) = L_{WP}+ \lambda L_{Dist}
\end{equation}
where $\varphi$ corresponds to the the parameters of the state encoder, $\theta$ represents the parameters of the diffusion policy. The hyperparameter $\lambda=10^{-4}$ controls the relative weight of the temporal distance loss.

After DAgger learning, we fine-tune the policy using the augmented demonstrations collected during interaction, while retaining the same hyperparameter settings to obtain the final trained policy.

\section{Experiment}
\subsection{Experiment Setting}
\subsubsection{Dataset}

We conducted experiments on the R2R-CE dataset which extends the high-level trajectories of R2R dataset\cite{anderson2018vision} into continuous environments using the Habitat Simulator \cite{savva2019habitat}, allowing agents to navigate smoothly across the ground surface. 
The R2R-CE dataset comprises 16,833 instruction-trajectory pairs, based on 5,611 human-annotated navigation trajectories, each paired with three different English instructions. It spans 90 environments, with 61 used for training, 11 for validation, and 18 for testing. 
The dataset includes distinct validation and test splits: the Val\_seen split contains novel trajectories within the same environments as the training set, while the Val\_unseen split introduces new trajectories in previously unseen environments. The Test set provides only the goal positions in novel environments.

To assess the ability of the proposed method to model multi-modal action distributions, we divide the dataset into several subsets based on environmental characteristics, allowing us to evaluate the model performance under different scene conditions. Specifically, we categorize the dataset into three types of environments:
(1) Open Area - the agent primarily navigates through wide, unobstructed spaces with minimal obstacles;
(2) Narrow Space - the agent operates in tight, confined environments with nearby obstacles, which often lead to navigation failures due to unreachable targets;
(3) Stairs - scenes that contain staircases, which frequently lead to incorrect waypoint predictions.
For each environment category, we randomly select one scene and train a model exclusively on its corresponding data. We then report the experimental results to evaluate the method's performance in that specific environment.

\subsubsection{Evaluation Metrics}
Following previous works \cite{an2024etpnav, hong2021vln}, we adopt the following navigation metrics. Trajectory Length (TL): average path length in meters;  Navigation Error (NE): average distance between the agent’s final position and the goal location; Success Rate (SR): the ratio of paths with NE less than 3 meters; Oracle SR (OSR): SR given oracle stop policy; SR penalized by Path Length (SPL); Collision Rate (CR): the average number of collisions per navigation episode.

\subsubsection{Model Configuration}
For agent's state encoding, we use a ViTB/32 pre-trained in CLIP \cite{radford2021learning} to encode observations, and a pre-trained ResNet-50 \cite{he2016deep} to encode depth images following. The text encoder that we used is LXBERT\cite{tan2019lxmert}, and all parameters of the visual encoder and the text encoder are frozen.
We set the number of panorama encoder and cross-modal transformer layers following \cite{an2024etpnav}, and initialize the model with pretrained LXMERT on the R2R-CE dataset in the default policy training. As mentioned in \ref{Training Details}, we set the interval $n=2$ to sample the waypoints of the expert trajectories and generate sparse training trajectories.

For action prediction, we adopt the same diffusion policy architecture as NoMaD\cite{sridhar2024nomad}, using a 1D conditional U-Net with 15 convolutional layers as the noise prediction network$\epsilon_\theta$. The diffusion policy is trained using the Square Cosine Noise Scheduler\cite{nichol2021improved} with K = 10 denoising steps. We employ the AdamW optimizer with a learning rate of $10^{-4}$ and train DifNav for 100 epochs with a batch size of 256.

\subsection{Comparison with State-Of-The-Art Methods}

\begin{table*}[!t]
\caption{Experimental results evaluated on the R2R-CE dataset\label{table:main}}
\centering
\setlength{\tabcolsep}{2mm}{
\begin{tabular}{cc|c|cccc|cccc|cccc}
\hline
& \multicolumn{1}{c|}{Scene} & \multirow{2}*{Framework} & \multicolumn{4}{c|}{Open Area} & \multicolumn{4}{c|}{Narrow Space} & \multicolumn{4}{c}{Stairs} \\
\cline{1-2} \cline{4-15}
 & Methods  &   & NE$\downarrow$ & OSR$\uparrow$ & SR$\uparrow$ & SRL$\uparrow$  & NE$\downarrow$ & OSR$\uparrow$ & SR$\uparrow$ & SRL$\uparrow$  & NE$\downarrow$ & OSR$\uparrow$ & SR$\uparrow$ & SRL$\uparrow$ \\
\hline
1 & GridMM  & \multirow{4}*{Two-Stage} & 3.5 & 81.0 & 71.4 & 65.5    & 2.9 & 80 & 60 & 57.9        & 4.3 & 66.7 & 66.7 & 61.8 \\
2 & ETPNav  &  & 3.3 & 81.0 & 76.2 & 72.6    & 1.7 & 80.0 & 73.3 & 68.7    & 5.0 & 66.7 & 61.9 & 60.9 \\
3 & BEVBert &  & 3.3 & 66.7 & 66.7 & 60.0    & 2.2 & 93.3 & 80 & 76.4      & 2.5 & 85.7 & 85.7 & 79.2 \\
4 & HNR-VLN &  & 3.7 & 81.0 & 81.0 & 70.8    & 2.0 & 86.7 & 73.3 & 63.9    & 5.3 & 66.7 & 61.9 & 59.1 \\
\hline
5 & CMA+PM+DA+Aug & \multirow{2}*{End-to-End} & 7.6 & 42.9 & 42.9 & 37.0    & 4.7 & 60 & 20 & 18.2        & 7.0 & 38.1 & 33.3 & 31.3 \\
6 & SEQ+DA & & 9.6 & 28.6 & 28.6 & 28.5                                     & 4.6 & 60 & 33.3 & 29.5      & 9.3 & 42.9 & 28.6 & 27.6 \\
\hline
7 & Waypoint Models & \multirow{4}*{End-to-End} & 5.1 & 42.9 & 36.8 & 36.8                           & 5.1 & 46.7 & 33.3 & 29.2   & 5.4 & 47.6 & 47.6 & 43.2 \\
8 & CM$^2$ & & 5.0 & 52.4 & 47.6 & 42.1           & 5.1 & 46.7 & 46.7 & 44.8    & 7.3 & 38.1 & 33.3 & 32.8 \\
9 & WS-MGMap & & 4.0 & 57.1 & 57.1 & 53.9                                  & 3.1 & 80 & 66.7 & 64.0    & 4.5 & 60 & 53.3 & 50.1 \\
10 & DifNav(Ours) & & \bf{2.2} & \bf{90.5} & \bf{90.5} & \bf{89.7}          & \bf{1.5} & \bf{93.3} & \bf{93.3} & \bf{91}       
 & \bf{1.5} & \bf{100} & \bf{90.5} & \bf{85.6} \\
\hline

\end{tabular}}
\end{table*}

We compare our method against three categories of VLN-CE models: (1) two-stage waypoint-based models, (2) end-to-end models that operate within a predefined discrete action space, and (3) end-to-end models that directly predict actions in continuous navigation space.

The results can be found in Table \ref{table:main}. Compared to models in the same category, DifNav significantly outperforms the second-best model, WS-MGMap, with average improvements of 28 in OSR, 32 in SR, and 32 in SPL. This significant performance gain can be attributed to the diffusion policy's strong capability in approximating multi-modal action distributions in long-horizon navigation tasks, as well as the use of DAgger to mitigate compounding errors during end-to-end training via behavior cloning.
Compared to models that operate in the low-level action space, DifNav achieves substantial improvements over the best-performing method, CMA, with average gains of 47 in OSR, 59 in SR, and 59 in SPL. This improvement is primarily attributed to the direct prediction of actions in 3D space, which significantly reduces the sequence length required to model long-horizon navigation behaviors.
Compared to models with a two-stage framework, DifNav outperforms the state-of-the-art method BEVBert by an average of 12 OSR, 13 SR, and 18 SPL, demonstrating the effectiveness of the proposed approach.
Notably, despite not relying on a trained waypoint predictor to reduce the sequence length for modeling, DifNav consistently outperforms two-stage models across all evaluated scenes.

The superior performance of DifNav can be attributed to its end-to-end architecture, which enables direct optimization of the navigation task. This design effectively mitigates the issue of error propagation commonly observed in two-stage models, where inaccurate waypoint predictions in the first stage can negatively impact downstream planning. Furthermore, by modeling multi-modal action distributions with a diffusion policy, DifNav is better able to capture the behavioral diversity exhibited in expert demonstrations and online augmented data during DAgger training.

Another important observation is that our method demonstrates consistent performance across different environments, effectively mitigating the performance fluctuations commonly observed in models that rely on waypoint predictors, which are sensitive to environmental variations.

\subsection{Ablation Study}

In this section, we present detailed ablation studies to assess the impact of specific design choices on DifNav, including historical information representation, waypoint spacing in demonstration trajectories, navigation completion detection strategies, and the effectiveness of incorporating DAgger training.
\subsubsection{Representations of Historical Information}

We compare different strategies for modeling the history trajectory as the representation of historical information: (1) using the three or five most recent observations in history trajectory; (2) using a sparse history trajectory obtained by sub-sampling every fourth waypoint; and (3) using only the current observation, thereby excluding any historical information.
\begin{table}[!ht]
\caption{Comparison of different representation for history modeling\label{table:Aba-his}}
\centering
\begin{tabular}{c|ccccc}
\hline
 History Form & NE$\downarrow$ & OSR$\uparrow$ & SR$\uparrow$ & SRL$\uparrow$ & CR$\downarrow$\\
\hline
Current Observation  & 4.4 & 49.5 & 49.5 & 46.6 & \bf{7.1} \\
Sparse History Trajectory  & 1.9 & 90.8 & 89.2 & 86.4 & 19.8 \\
5 Latest Observations  &  2.6 & 82.2 & 74.3 & 68.7 & 9.2 \\
3 Latest Observations  & \bf{1.7} & \bf{94.4} & \bf{91.4} & \bf{88.8} & 8.7 \\

\hline
\end{tabular}
\end{table}

The results, shown in Table \ref{table:Aba-his}, indicate that using only the current observation achieves the best obstacle avoidance performance, i.e., the lowest collision rate of 7.1 collisions per episode. This outcome is unsurprising, as current observation alone often provides sufficient information for effective short-term obstacle avoidance.
Interestingly, using the three most recent observations outperforms both the five-observation and sparse-history settings, despite the latter incorporating more historical information. One possible explanation is that excessive historical information may dilute the model's attention and hinder its ability to focus on the most relevant current observations for effective decision-making.
Moreover, longer history trajectories may include misleading information, such as the agent becoming stuck or repeatedly turning in place. In contrast, limiting the input to the three most recent observations helps “refresh” the state representation, filtering out such noise while still retaining sufficient context for grounding the navigation instruction.

\subsubsection{Waypoint Spacing in Demonstration Trajectories}

As discussed above, the waypoint spacing in demonstration trajectories influences the action generation space of DifNav. To investigate the optimal spacing configuration, we evaluate policies trained with demonstration trajectories of different waypoint spacings: 0.25, 0.5, and 1 meter. We expect the policy to produce outputs that match the spacing of the corresponding demonstrations.
\begin{table}[!ht]
\caption{Comparison of different Waypoint Spacing in Demonstration Trajectories\label{table:Aba-ws}}
\centering
\begin{tabular}{c|ccccc}
\hline
Step Size & NE$\downarrow$ & OSR$\uparrow$ & SR$\uparrow$ & SRL$\uparrow$ & TL\\
\hline
waypoint\_spacing=0.25m  & 2.8 & 73.0 & 71.4 & 68.8 & 8.0 \\
waypoin\_spacing=0.50m  & \bf{1.7} & \bf{94.4} & \bf{91.4} & \bf{88.8} & 9.8 \\
waypoint\_spacing=1.0m  & 2.3 & 86.0 & 78.1 & 62.3 & 12.6 \\
\hline
\end{tabular}
\end{table}

The results in Table \ref{table:Aba-ws} suggest that training the model with demonstration trajectories that have either excessively small or large waypoint spacings leads to degraded navigation performance. This can be attributed to two factors: overly small spacings increase the action horizon, making modeling more difficult, while overly large spacings make single-step action prediction more challenging due to the sparsity of supervision.
Furthermore, as the waypoint spacing increases, the resulting navigation trajectories become longer, which can be explained by the expanded action generation space available to the policy.

\subsubsection{Navigation Completion Detection Network}
We explore the most suitable representation for determining whether the agent has reached the navigation goal.
In the first setting, we use a 3-layer MLP to predict the normalized distance to the goal at each decision step, and ask the agent to stop when this distance falls below a pre-defined threshold.
In the second setting, a 2-layer MLP followed by a regularization layer is used to directly predict whether the agent has reached the goal.
In the third approach, we address the severe class imbalance inherent in long-horizon navigation tasks by applying a weighting factor to the stop label during training.
\begin{table}[!ht]
\caption{Comparison of different stop strategy\label{table:Aba-stop}}
\centering
\begin{tabular}{c|ccccc}
\hline
 Stop Form & NE$\downarrow$ & OSR$\uparrow$ & SR$\uparrow$ & SRL$\uparrow$ & TL\\
\hline
Normalized Distance  & \bf{1.7} & \bf{94.4} & \bf{91.4} & \bf{88.8} & 9.8 \\
Classification Stop  & 2.3 & 86.3 & 80 & 77.6 & 9.9 \\
Weighted Classification Stop  & 5.3 & 17.5 & 17.5 & 17.5 & \bf{4.6} \\
\hline
\end{tabular}
\end{table}

The results, shown in Table \ref{table:Aba-stop}, indicate that using normalized distance to determine whether the agent should stop outperforms the classification-based approach. This is expected, because asking the agent to decide when to stop is an extremely class-imbalanced task in long-horizon navigation. In contrast, predicting the normalized distance serves as a form of progress monitoring, enabling the agent to assess how far it has advanced in following the instruction.
Applying a weighting factor to the loss of the minority class is a common strategy to address class imbalance. However, in our case, this approach severely degrades the policy’s navigation ability and results in abnormally short trajectories, suggesting that the agent learns to stop prematurely during navigation.

\subsubsection{DAgger-Boosted Online Training}
\begin{table}[!ht]
\caption{Evaluation of the effectiveness of DAgger training\label{table:Aba-DAgger}}
\centering
\begin{tabular}{c|ccccc}
\hline
DAgger Training & NE$\downarrow$ & OSR$\uparrow$ & SR$\uparrow$ & SRL$\uparrow$ & CR$\downarrow$\\
\hline
w/o DAgger  & 5.2 & 59.3 & 21.0 & 15.9 & 238.2 \\
w DAgger($\alpha=0.10$)  & \bf{2.2} & \bf{87.6} & \bf{83.3} & 71.0 & \bf{14.0} \\
w DAgger($\alpha=0.25$)  & 2.7 & 78.4 & 76.2 & \bf{75.4} & 14.66 \\
w DAgger($\alpha=0.50$)  & 2.5 & 81.9 & 77.1 & 74.4 & 17.12 \\
w DAgger($\alpha=0.75$)  & 3.2 & 68.6 & 63.8 & 61.0 & 16.16 \\
\hline
\end{tabular}
\end{table}
Table \ref{table:Aba-DAgger} presents the performance of the proposed method using DAgger training under varying probability thresholds $\alpha$ for querying expert intervention.
A higher value of $\alpha$ results in the agent’s rollout trajectories closely following expert demonstrations, while a lower value of $\alpha$ grants the agent greater freedom to explore the environment.
\begin{figure*}[!ht]
\centering
\includegraphics[width=7in]{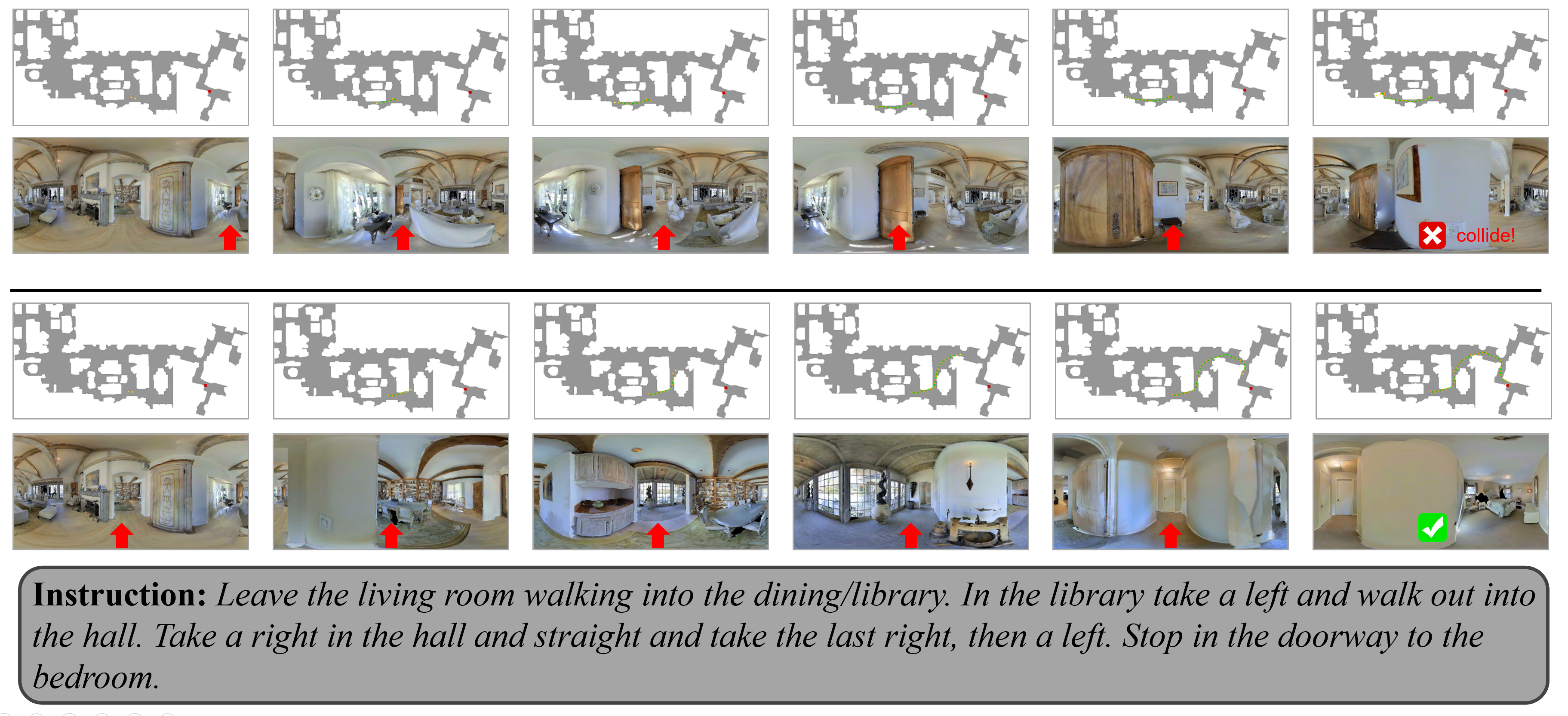}
\caption{Qualitative analysis of the effectiveness of DAgger-boosted online training in addressing compounding errors.
Top: Without DAgger training, the policy fails to complete the navigation task, even when following training trajectories. Bottom: After applying DAgger, the agent navigates correctly and smoothly to the goal.}
\label{fig:DAgger}
\end{figure*}

\begin{figure*}[!ht]
\centering
\includegraphics[width=7in]{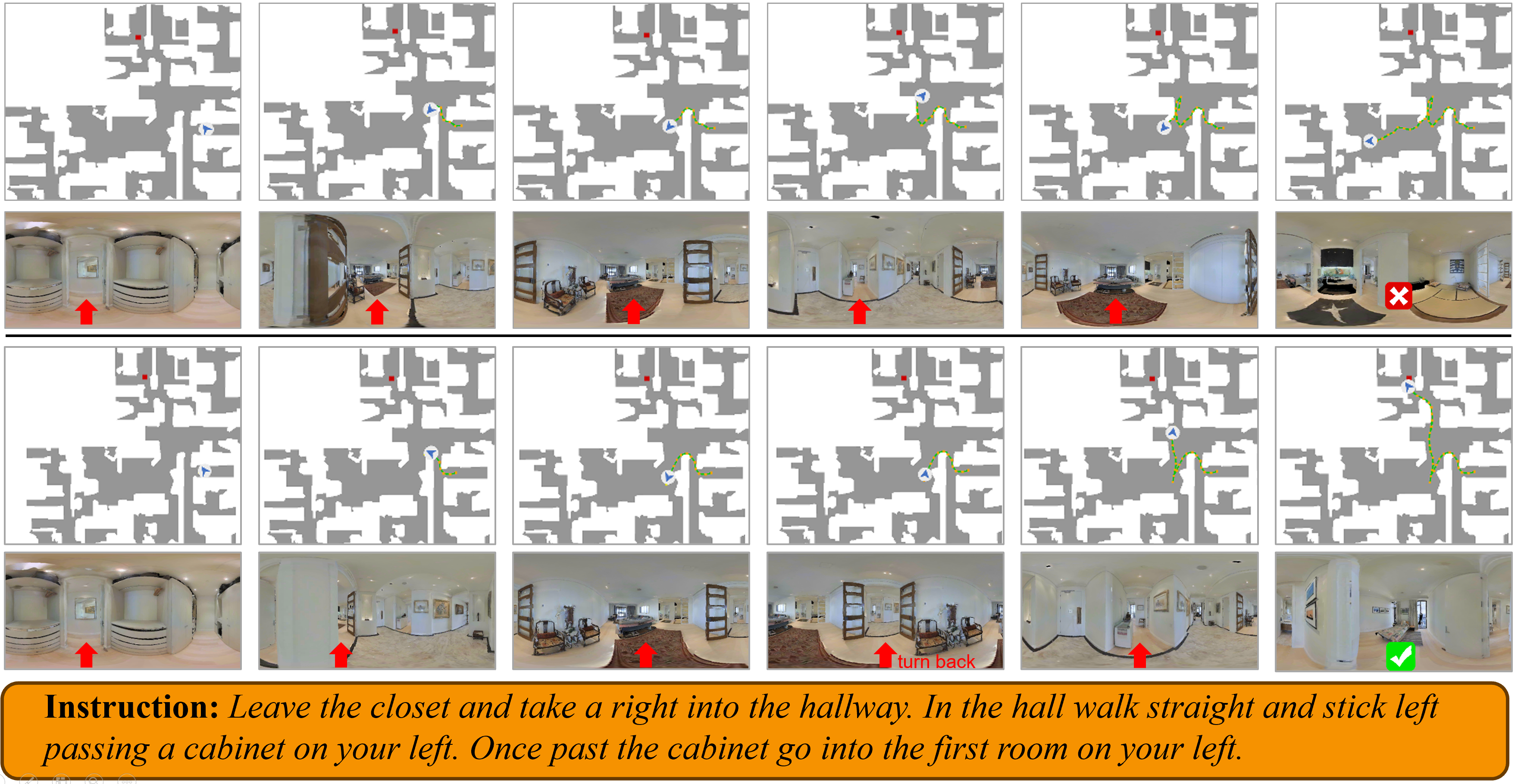}
\caption{Qualitative analysis of navigation error correction. 
Top: The policy trained only on expert demonstrations fails to revise its plan when navigation errors occurring. Bottom: The policy trained with augmented data learns to revise its plan upon encountering error states.}
\label{fig:recover}
\end{figure*}
\begin{figure*}[!ht]
\centering
\includegraphics[width=7in]{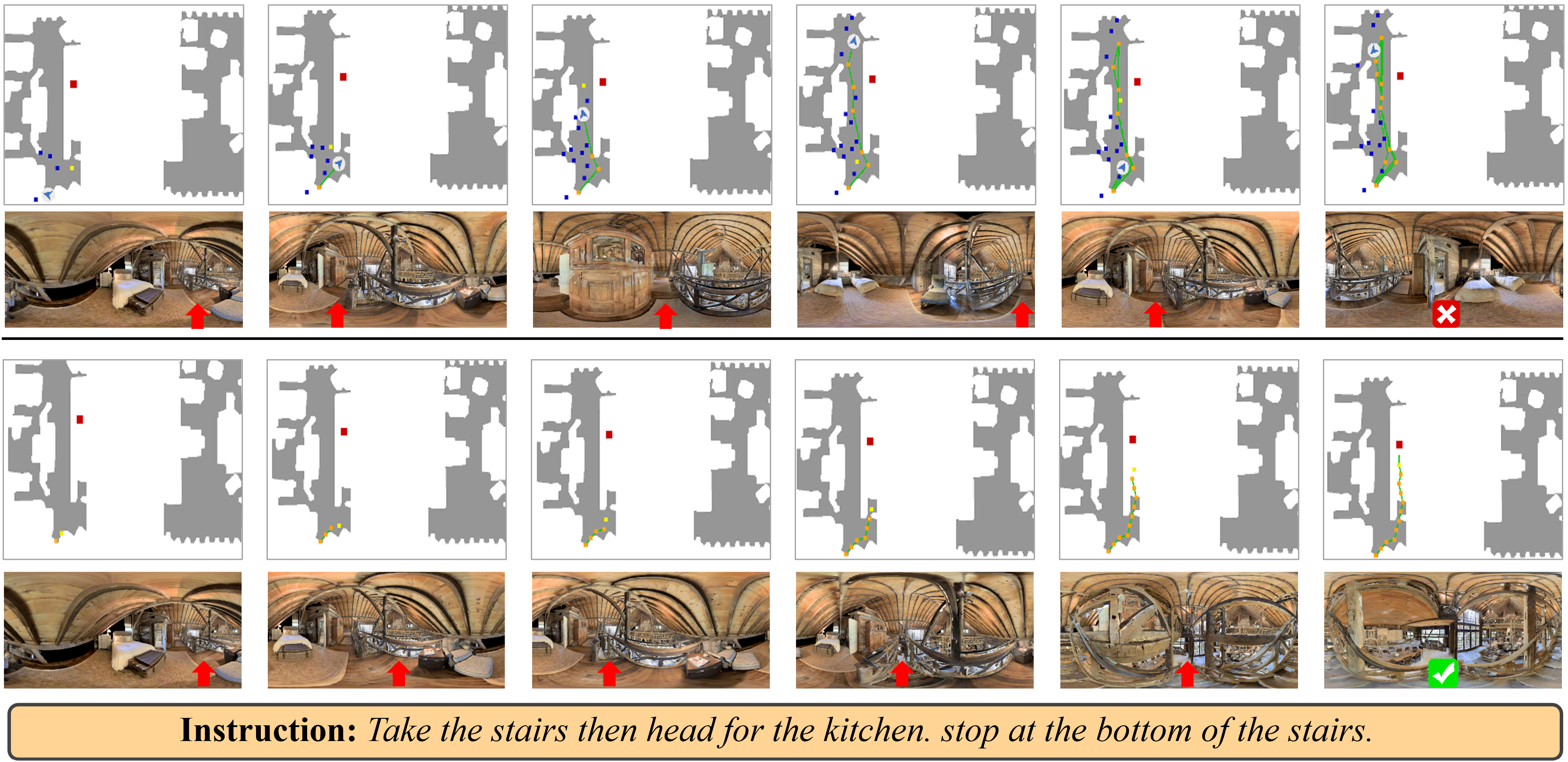}
\caption{Comparison between two-stage waypoint-based methods and DifNav.
Top: The two-stage method fails when the waypoint predictor cannot generate waypoints that lead to the goal (the bottom of the stairs). Bottom: DifNav directly models a multi-modal action distribution and samples actions that successfully guide the agent to the goal.}
\label{fig:wp-DDN}
\end{figure*}
The poor performance of the policy without online training highlights the severity of compounding errors caused by the distributional shift between training and testing, as well as the limited availability of expert trajectories during BC.
After DAgger-boosted online training, this issue is significantly mitigated, and the policy performance improves further in the lower $\alpha$ setting.
A reasonable explanation is that during DAgger training, the agent learns to recover from error states by querying expert interventions when it deviates from the correct path. Once equipped with this skill, the agent can "turn around" when straying from the navigation target or encountering obstacles. Lower $\alpha$ setting allows the agent to explore more diverse situations requiring corrective actions, enabling the policy to learn from a broader range of challenging scenarios, which ultimately leads to better overall performance.
However, a lower value of $\alpha$ in DAgger training leads to lower SPL and longer output trajectories. A possible explanation is that, with reduced expert guidance, the online-collected trajectories deviate more from the expert demonstrations. As a result, the policy tends to produce suboptimal actions initially and subsequently corrects them, which contributes to increased trajectory lengths.

\subsection{Qualitative Analysis}
In this section, we provide a qualitative analysis from the perspectives of addressing compounding errors, error recovery, and policy navigability.
\subsubsection{Addressing the Severe Compounding Errors}
The default policy in our method is initially trained via BC using demonstration trajectories from benchmark datasets. This means that the policy is only exposed to a limited subset of states within the 3D continuous environment.
This constraint does not significantly impact two-stage waypoint-based methods, since the exploration space is effectively defined once the waypoint predictor is well trained and the starting location is given. However, end-to-end approaches, which directly predict actions as a generative task in continuous 3D space, will suffer a severe problem - compounding errors. 
Although the model tends to output reasonable navigation actions in the initial steps, slight deviations from the demonstration action can still occur. These per-step errors gradually accumulate, eventually causing the policy to drift off-distribution over long horizons.
As illustrated in Fig. \ref{fig:DAgger}, DifNav trained solely on demonstration trajectories fails to complete navigation tasks even in the training set. In contrast, our DAgger-boosted online learning strategy collects additional online navigation trajectories to augment the training distribution. By exposing the model to states near the expert trajectory, this strategy enhances the robustness of the policy and mitigates compounding errors substantially during navigation.

\subsubsection{Recover from Incorrect States}
In end-to-end model training via BC, the model cannot learn to replan when a navigation error occurs, as expert trajectories typically do not contain examples of error states or recovery behavior. As illustrated in Fig. \ref{fig:recover} (top), when a navigation error occurs, the model becomes confused and continues toward an incorrect destination.
To overcome this limitation, we apply the DAgger algorithm, which allows the policy to execute its own actions with a certain probability, while expert intervention is triggered otherwise. This process enables the collection of augmented trajectories that include corrective supervision. For example, a previous step may involve a suboptimal action generated by the policy, and the following step uses an expert action to recover from the error. As shown in Fig. \ref{fig:recover} (down), the policy learns to turn back when a navigation failure occurs, indicating improved robustness and recovery capability.

\subsubsection{Two-Stage Waypoint-Based Approach vs DifNav}
A potential performance bottleneck in two-stage waypoint-based approaches lies in the sub-optimality of the waypoint predictor. If the predictor fails to generate suitable waypoints that lead to the navigation goal, the agent cannot successfully complete the task. As illustrated in Fig. \ref{fig:wp-DDN} (top), the two-stage method ETPNav is constrained to roll out within the set of waypoints provided by the predictor. However, in this example, the waypoint predictor does not generate any waypoints connecting to the stairs, which serve as the navigation goal, resulting in the failure of the task.
In contrast, DifNav directly models a multi-modal action distribution in continuous navigation space and samples the next action from this distribution. This formulation eliminates the reliance on intermediate waypoints and, in principle, allows the policy to navigate to any location in the environment.

\section{Conclusion}
In this work, we present DifNav, the first attempt to leverage a diffusion policy for modeling the multi-modal action distribution in vision-and-language navigation. Our approach adopts an end-to-end training framework to address the global suboptimality issue inherent in two-stage waypoint-based strategies. By using the diffusion policy to capture the multi-modal action distribution, our policy can predict multiple correct actions resulting from ambiguous instructions or the existence of several equivalent paths. To mitigate the compounding errors commonly encountered in end-to-end training with behavioral cloning, we incorporate DAgger to augment the training data distribution for fine-tuning, thereby improving policy robustness. Extensive experiments demonstrate that DifNav more effectively models multi-modal action distributions and achieves SOTA navigation performance across different environments.

\bibliographystyle{IEEEtran}
\bibliography{references}

\end{document}